\documentclass{melba}

\usepackage{mwe} 

\usepackage{amsmath,amsfonts}

\melbaid{2025:NNN}  
\doi{10.59275/j.melba.2024-AAAA}
\melbaauthors{H. Jiang, S. Xie and Z. Wan}  
\email{wanzhy@shanghaitech.edu.cn}
\volume{2}
\firstpageno{1337}  
\melbayear{2025}  
\datesubmitted{yyyy-m1-d1}  
\datepublished{yyyy-m2-d2}  

\melbaspecialissue{Medical Imaging with Deep Learning (MIDL) 2020}
\melbaspecialissueeditors{Marleen de Bruijne, Tal Arbel, Ismail Ben Ayed, Hervé Lombaert}

\ShortHeadings{A DICOM Image De-identification Algorithm in the MIDI-B Challenge}{H. Jiang, S. Xie and Z. Wan}

\title{A DICOM Image De-identification Algorithm in the MIDI-B Challenge}

\author{
	\firstname Hongzhu \surname  Jiang\aff{1},
    \firstname Sihan \surname  Xie\aff{1},
	\firstname  Zhiyu \surname Wan\aff{1,2}\orcid{0000-0003-3752-5778}
}
\affiliations{
	\num 1 \addr ShanghaiTech University, Shanghai 201210, China \\
	\num 2 \addr Vanderbilt University Medical Center, Nashville TN 37203, USA 
}

\abstract{
	Image de-identification is essential for the public sharing of medical images, particularly in the widely used Digital Imaging and Communications in Medicine (DICOM) format as required by various regulations and standards, including Health Insurance Portability and Accountability Act (HIPAA) privacy rules, the DICOM PS3.15 standard, and best practices recommended by the Cancer Imaging Archive (TCIA). The Medical Image De-Identification Benchmark (MIDI-B) Challenge at the 27th International Conference on Medical Image Computing and Computer Assisted Intervention (MICCAI 2024) was organized to evaluate rule-based DICOM image de-identification algorithms with a large dataset of clinical DICOM images. In this report, we explore the critical challenges of de-identifying DICOM images, emphasize the importance of removing personally identifiable information (PII) to protect patient privacy while ensuring the continued utility of medical data for research, diagnostics, and treatment, and provide a comprehensive overview of the standards and regulations that govern this process. Additionally, we detail the de-identification methods we applied — such as pixel masking, date shifting, date hashing, text recognition, text replacement, and text removal — to process datasets during the test phase in strict compliance with these standards. According to the final leaderboard of the MIDI-B challenge, the latest version of our solution algorithm correctly executed 99.92\% of the required actions and ranked 2nd out of 10 teams that completed the challenge (from a total of 22 registered teams). Finally, we conducted a thorough analysis of the resulting statistics and discussed the limitations of current approaches and potential avenues for future improvement. }


\begin{document}
\sloppy
\twocolumn[\maketitle]

\section{Introduction}
        \subsection{MIDI-B De-identification}
        \setlength{\parindent}{2em}
        In the era of digital healthcare, the processing and analysis of medical images are critical for diagnostics, treatment planning, and research \citep{aggarwal2021diagnostic}. One of the key challenges in this domain is the de-identification of medical images \citep{Chevrier2019, Moore2012}, which involves removing or obfuscating personally identifiable information (PII) to protect patient privacy while preserving data utility. This process is essential for ensuring compliance with privacy regulations like General Data Protection Regulation (GDPR) \citep{voigt2017eu} and Health Insurance Portability and Accountability Act (HIPAA) \citep{annas2003hipaa}, and for promoting the responsible sharing of medical data.       
        
        Medical images, particularly in the widely used Digital Imaging and Communications in Medicine (DICOM) format \citep{bidgood1992}, contain not only the image data but also sensitive metadata, such as patient names and birthdates. While this metadata is invaluable for  clinical workflows, it poses significant privacy risks if not properly de-identified \citep{moore2015identification}. Therefore, effective de-identification of DICOM images is crucial for safely managing and sharing medical imaging data. 
        
        To address these challenges, the Medical Image De-Identification Benchmark (MIDI-B) challenge evaluates rule-based DICOM de-identification algorithms using a diverse set of standardized clinical images with synthetic identifiers \citep{kushida2012}. This competition aims to advance automated de-identification methods that maintain both privacy and data utility, thereby supporting the secure sharing of medical research data and fostering innovation in privacy-preserving technologies.
        
   \subsection{    De-identification Standards}
         In the context of the MIDI-B challenge, de-identification (deID) refers to adherence to the US HIPAA Privacy Rule safe harbor method, DICOM Standard PS3.15 (Attribute Confidentiality Profile), and Best Practices described in the TCIA Submission Overview Page.
        \paragraph{HIPAA Privacy Rule Safe Harbor method.}The HIPAA Privacy Rule is a U.S. regulation designed to protect the privacy of individuals' health information. By removing or obscuring identifiable information, it ensures that medical data does not reveal personal identities, thereby supporting the lawful use and sharing of data.

        The Safe Harbor Provision is part of the HIPAA Privacy Rule and provides a method for ensuring that Protected Health Information (PHI) is appropriately de-identified. This involves removing or altering personal identifiers so that data cannot be traced back to specific individuals. According to HIPAA, de-identification requires the removal of 18 types of information, including names, addresses, birthdates, Social Security numbers, medical record numbers, and insurance policy numbers. Retaining this information could lead to the identification of specific individuals.
        \begin{table}[h]
            \centering
            \caption{List of HIPAA identifiers.}
            \begin{tabular}{p{0.06\columnwidth} p{0.85\columnwidth}}
            \textbf{No.} &\textbf{Identifier Type}\\
            \hline
            1 & Name/Initials \\
            2 & Street address, city, county, precinct code and equivalent geocodes for ZIP-3 when population is of size < 20,000 people \\
            3 & Dates (indicative of a time period smaller than 1 year) and all ages over 89\\
            4 & Telephone Numbers \\
            5 & Fax Numbers \\
            6 & Electronic Mail Address \\
            7 & Social Security Number\\
            8 & Medical Record Number\\
            9 & Health Plan ID Number\\
            10 & Account Number\\
            11 & Certificate / License Number\\
            12 &  Vehicle identifiers and serial numbers, including license plate numbers\\
            13 & Device Identifiers and serial numbers \\
            14 & Web addresses (URLs) \\
            15 & Internet IP Addresses \\
            16 & Biometric identifiers, including finger and voice prints\\
            17 &  Full face photographic images and any comparable images \\
            18 & Any other unique identifying number, characteristic, or code
            \end{tabular}
        \end{table}
        \paragraph{DICOM Standard PS3.15.} DICOM Standard PS3.15 (Attribute Confidentiality Profile) is a component of the DICOM standard that focuses on the confidentiality of attributes in medical imaging data \citep{dicom2016}. This standard defines how to protect sensitive information in DICOM data to ensure patient privacy and data security \citep{tanabe2018}. 
        
        It specifies requirements for safeguarding specific attributes, such as patient names and birthdates, from unauthorized access. It also outlines methods for handling and filtering sensitive attributes in DICOM images to prevent exposure during data sharing or transmission and provides de-identification procedures to remove or obscure patient identity information.
        \paragraph{The Cancer Imaging Archive (TCIA).}The TCIA Submission Overview Page outlines best practices for submitting medical imaging data to the Cancer Imaging Archive (TCIA) \citep{clark2013}. It provides clear guidelines to ensure data quality and privacy protection, facilitating the submission and utilization of high-quality data while meeting regulatory requirements.

        In terms of de-identification, TCIA ensures that all personally identifiable information (PII) is removed from images and associated metadata, including patient names, IDs, and other demographic details. It employs standardized de-identification procedures to comply with privacy regulations such as HIPAA and thoroughly reviews data to ensure that no identifiable information remains.

        These three de-identification standards effectively protect patient privacy while ensuring the research value and security of the data.
\section{Methods}
    We implemented two categories of de-identification methods: 1) simple de-identification and 2) pseudonymization. We define simple de-identification as the removal of real patient identifiers, while pseudonymization involves replacing identifiers with a pseudonym that is unique to the individual and known within a specified context but not linked to the individual in the external world. For simple de-identification, we employed methods such as text recognition, text removal, and pixel masking. In contrast, for pseudonymization, we used techniques including date shifting, hashing, text recognition, and text replacement. Our de-identification methods are open-source and are available at \url{https://github.com/zhywan/midi-b-challenge-2024}.
    \subsection{Simple De-identification}
        \paragraph{Pixels Masking.}We utilized the Presidio \citep{presidio} \citep{kotevski2022}, a data protection and de-identification soft- ware  development kit  from Microsoft to remove pixels containing  sensitive information. Presidio can aid in the proper management and governance of sensitive data and offers rapid identification and anonymization modules for processing text and images. Additionally, it consists of three major functional modules: Analyzer, Anonymizer, and Image Redactor. The Analyzer is responsible for scanning text-based data to identify sensitive information. It includes predefined recognizers and can be extended with custom recognizers. The Anonymizer is focused on desensitizing detected sensitive entities. It replaces detected PII with anonymized values using operators such as substitution, masking, and ciphering. The Image Redactor uses OCR technology to identify and desensitize sensitive information within images. Presidio allowed us to use an analyzer to detect sensitive data and identify the pixel positions of these data in the original DICOM image and cover these pixel areas with color blocks.

        In our implementation, we used the Document Intelligence OCR  \citep{satapathi2024chapter6} provided by Microsoft Azure to identify text in pixel data. Additionally, we created allow lists (i.e., whitelists) and deny lists (i.e., blacklists) for this specific task, based on the validation dataset, with custom recognizers. The flow chart of the pixels masking of our approach is shown in Figure 1.We also fixed bugs in the original Presidio packages related to redaction colors and standardized the redaction color for consistency. We redacted the pixels for all DICOM files  in the  first round  and then processed the metadata for each file in the second round.

        Examples of the final implementation of the pixels masking action is depicted in Figure 2, where the left of the figure represents the DICOM file before the action, and the right of the figure shows the DICOM file after the action. And another example is shown in Figure 3.
        
        \begin{figure}[h]
        \centering
        \includegraphics[width=0.52\linewidth]{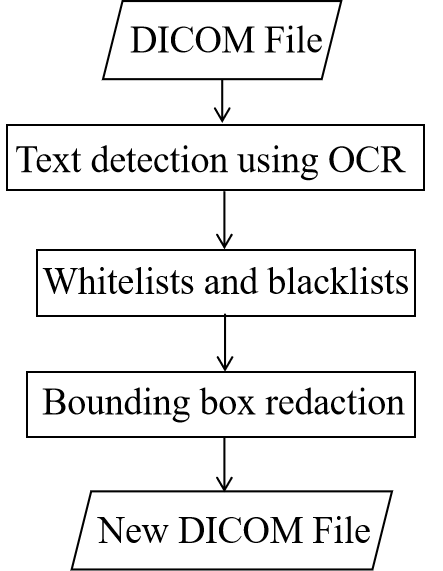}
        \caption{ The flow chat of the pixels masking in our approach.}
        \end{figure}
        
        \begin{figure}[h]
        \centering
        \includegraphics[width=0.8\linewidth]{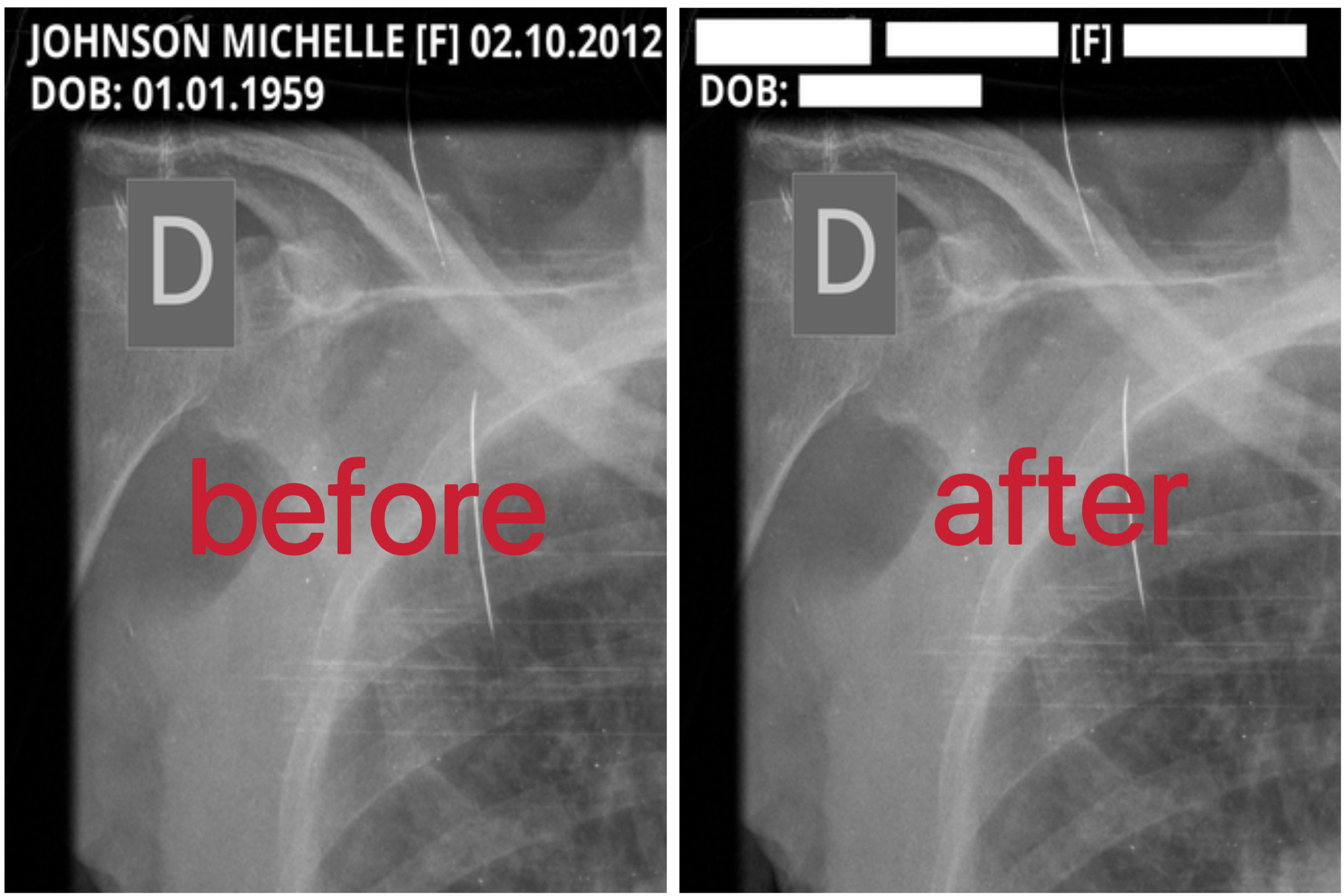}
        \caption{ An example of the final implementation of the pixel masking action.}
        \end{figure}
                \begin{figure}[h]
        \centering
        \includegraphics[width=0.65\linewidth]{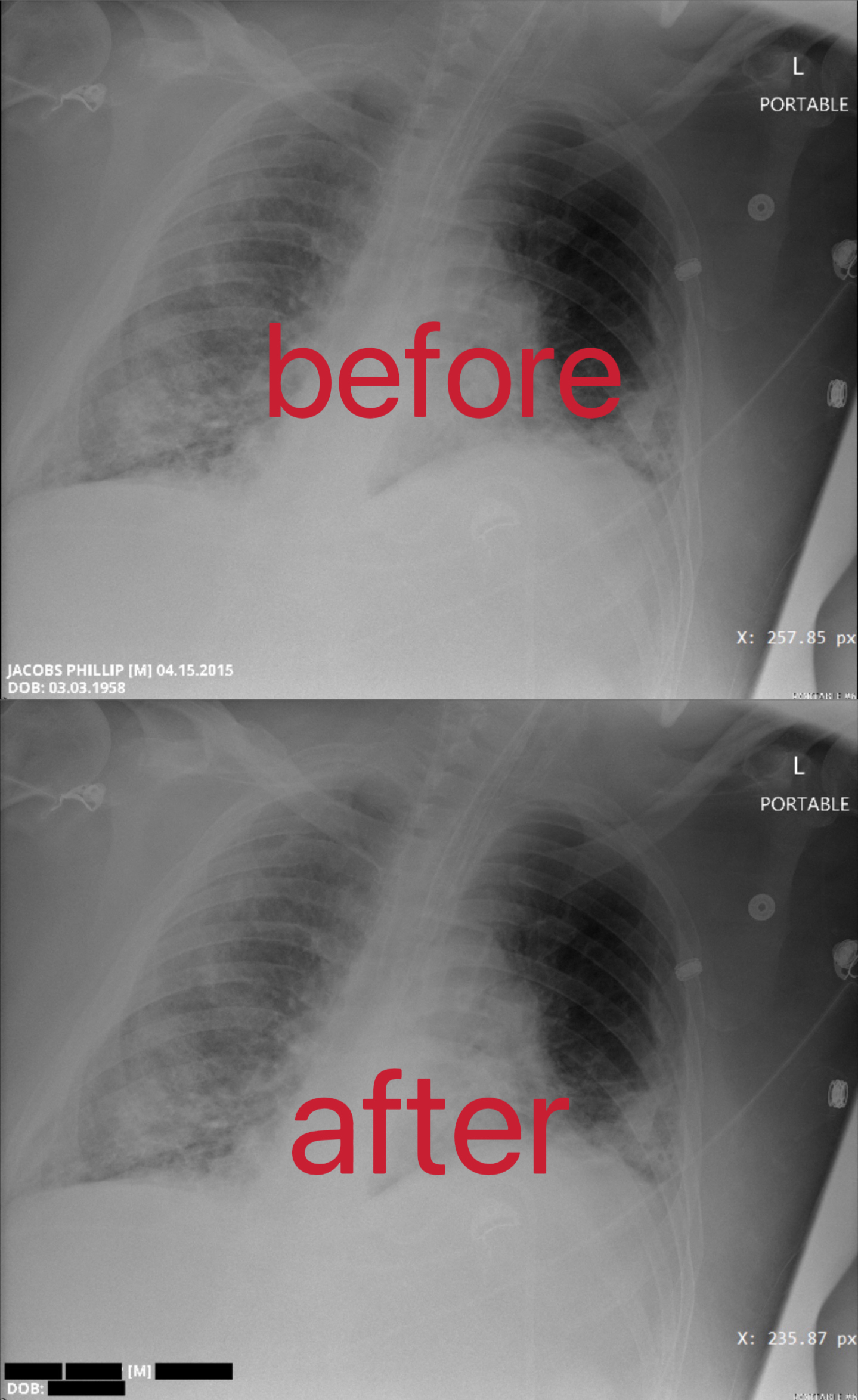}
        \caption{ Another example of the final implementation of the pixel masking action.}
        \end{figure}
        \paragraph{Text Removal.}The text specified should be removed from tag values. To protect patient privacy, it is necessary to remove all information that may identify a patient. Depending on the action code, specific tag values can be cleared, or the portions of the content containing personally identifiable information need to be removed.

        We removed the patient ID from the currently processed DICOM file by saving its value in a variable. There are 53 regular expressions designed in our algorithm to match and remove other sensitive information such as dates, phone numbers, and clinic names. Based on real-life experience, sensitive information such as abbreviations and names are identified by their relative position to prepositions such as “at” and “by”.

        \begin{table}[h]
            \centering
            \caption{Examples of tags from which the specified text needs to be removed from their values.}
            \begin{tabular}{p{0.18\columnwidth} p{0.475\columnwidth} p{0.08\columnwidth}p{0.1\columnwidth}}
            \textbf{Tag ID} &\textbf{Tag Description} & \textbf{Code}&\textbf{Action} \\
            \hline
            (0008,0080) & Institution Name & X & remove \\
            (0008,0081) & Institution Address & X & remove \\
            (0008,0094) & \parbox[t]{\linewidth}{Referring Physician's \\ Telephone Numbers} & X & remove \\
            (0008,0201) & Timezone Offset From UTC & X & remove
            \end{tabular}
        \end{table}
    \subsection{Pseudonymization}     
        \paragraph{Patient ID Replacement.}It is necessary to ensure that patient ID are replaced and are consistent with patient ID mapping. The patient ID mapping file contains two columns with the header names “id\_old” and “id\_new” . The “id\_old” column contains the patient ID before de-identification and “id\_new” is the patient ID after de-identification. Then we used a lookup function to replace both the patient ID and patient name with the new identifiers (i.e., “id\_new”). Table 3 shows tags that need to be updated according to the TCIA standards. Table 4 shows an example of a patient ID mapping file.

        \begin{table}[h]
            \centering
            \caption{Tags that need to be updated according to the TCIA standards.}
            \begin{tabular}{
            p{0.17\columnwidth}p{0.3\columnwidth}p{0.453\columnwidth}}
            \textbf{Tag ID} &\textbf{Tag Description} &\textbf{Action} \\
            \hline
            (0010,0010) & Patient Name & LOOKUP(PatientID, ptid) \\
            (0010,0020) & Patient ID & LOOKUP(this, ptid) \\
            \end{tabular}
        \end{table}

        \begin{table}[h]
            \centering
            \caption{An example of the patient ID mapping file.}
            \begin{tabular}{p{0.25 \columnwidth} p{0.2 \columnwidth} }
            \textbf{id\_old} &\textbf{id\_new}  \\
            \hline
            1059030585 & 0000001 \\
            1065842606 & 0000002 \\
            1097215536 & 0000003 \\
            1115564954 & 0000004 \\
            113575183 & 0000005
            \end{tabular}
        \end{table}
        \paragraph{UID Replacement.}DICOM makes extensive use of universal identifiers (UID) that could be used to identify a subject. We process the UID in the DICOM file through a hashing function.  First,  we determine a fixed-format prefix, e.g., uid\_root = ‘1.2.397.4.5. \{patient id\_new\}.8.117. ’. Then, we hash the UID to generate a unique hash value and keep the first 19 digits only. The final hashUID is generated by splicing the string \textit{uid\_root} in front of it as a prefix, where the “patient id\_new” in the prefix is a fixed-length numeric string. This method ensures that the \textit{id\_new} of UID for different patients are different, thus significantly reducing the possibility of \textit{collision} (i.e., two original values arehashed into the same value) in the hashing process.

        The mapping file contains two columns with the header names “id\_old” (the UID before de-identification) and “id\_new” (the UID after de-identification). The file provides a mapping of all UIDs pseudonymized during the participants' de-identification process to indicate the old UID and what it was transformed to. Table 5 shows an example of a UID mapping file. Table 6 shows examples of tags that the UID need to be updated according to the TCIA standards.

        \begin{table}[h]
            \centering
            \caption{An example of the UID mapping file.}
            \begin{tabular}{p{0.45 \columnwidth} p{0.45 \columnwidth} }
            \textbf{id\_old} &\textbf{id\_new}  \\
            \hline
            2.2.374.1.2.1964017.6.944.
            2103992807195684018 & 1.2.397.4.5.0000001.8.117.
            6881276565361048595 \\
            2.2.198.1.2.3201303.1.133.
            1364097273910791617 & 1.2.397.4.5.0000160.8.117.
            4832656936496443956 \\
            2.1.240.0.0.7462603.1.346.
            1406313923998980205 & 1.2.397.4.5.0000285.8.117.
            4677849556384631713 \\
            3.3.186.0.2.3750666.8.312.
            1032638300693915579 & 1.2.397.4.5.0000243.8.117.
            5463938882700346084 \\
            1.1.766.1.2.3936616.2.547.
            5374533990591717384& 1.2.397.4.5.0000246.8.117.
            7966161292648523333 \\
            \end{tabular}
        \end{table}

        \begin{table}[h]
            \centering
            \caption{An example of tags that the UID need to be updated according to the TCIA standards.}
            \begin{tabular}{p{0.18 \columnwidth} p{0.36 \columnwidth} p{0.35\columnwidth}}
            \textbf{Tag ID} &\textbf{Tag Description} &\textbf{Action} \\
            \hline
            (0008,0014) & Instance Creator UID & hashuid(@UIDROOT,
            this) \\
            (0008,0018) & SOP Instance UID & hashuid(@UIDROOT,
            this) \\
            (0008,1155) & \parbox[t]{\linewidth}{Referenced SOP \\ Instance UID}  & hashuid(@UIDROOT,
            this) \\
            (0008,3010) & Irradiation Event UID & hashuid(@UIDROOT,
            this) \\
            (0008,000D) & Study Instance UID & hashuid(@UIDROOT,
            this) \\
            \end{tabular}
        \end{table}

      \paragraph{Date Shifting.}To shift the date using the specified shift value, we generated a set of 322 different random numbers ranging from 1 to 365 to ensure that each patient's date offset was unique. We then subtracted the given number of offset days from the original date. This approach effectively de-identifies the date information by altering it in a way that prevents future date errors or anomalies.

There are four main formats for date values: 1) YYYYMMDD   2) YYYYMMDDHHMMSS   3) YYYYMMDDHHMMSS.FF  4) Unix timestamp. If the value includes time, we keep the time unchanged and only modify the date portion.

\begin{table}[h]
    \centering
    \caption{Example of tags that require the date to be shifted by a specified value.}
    \begin{tabular}{p{0.17 \columnwidth} p{0.3 \columnwidth} p{0.35\columnwidth}}
        \textbf{Tag ID} &\textbf{Tag Description} &\textbf{Action} \\
        \hline
        (0008,0012) & Instance Creation Date & incrementdate(this,
        @DATEINC) \\
        (0008,0020) & Study Date & incrementdate(this,
        @DATEINC) \\
        (0008,0021) & Series Date & incrementdate(this,
        @DATEINC) \\
        (0008,0022) & Acquisition Date & incrementdate(this,
        @DATEINC) \\
        (0008,0023) & Content Date & incrementdate(this,
        @DATEINC) \\
    \end{tabular}
\end{table}
\section{Results}
\subsection{ Dataset} The dataset we used during the validation phase contains 29,660 DICOM images with synthetic PHI and PII from 322 patients. Each patient folder contains one or more studies. Each study may have one or more series, and each series has one or more instances. There are 813 tags in the Data Set with non-null values, except for meta information. According to the Best Practices described in the TCIA Submission Overview Page, 44 tags need to be removed, 32 tags need to keep the original values unchanged, 2 tags need a lookup function, 17 tags need to be shifted by a hash function, and 10 tags need additional processing. Figure 4 shows the distribution of these action types.

\begin{figure}[h]
    \centering
    \includegraphics[width=1.02\linewidth]{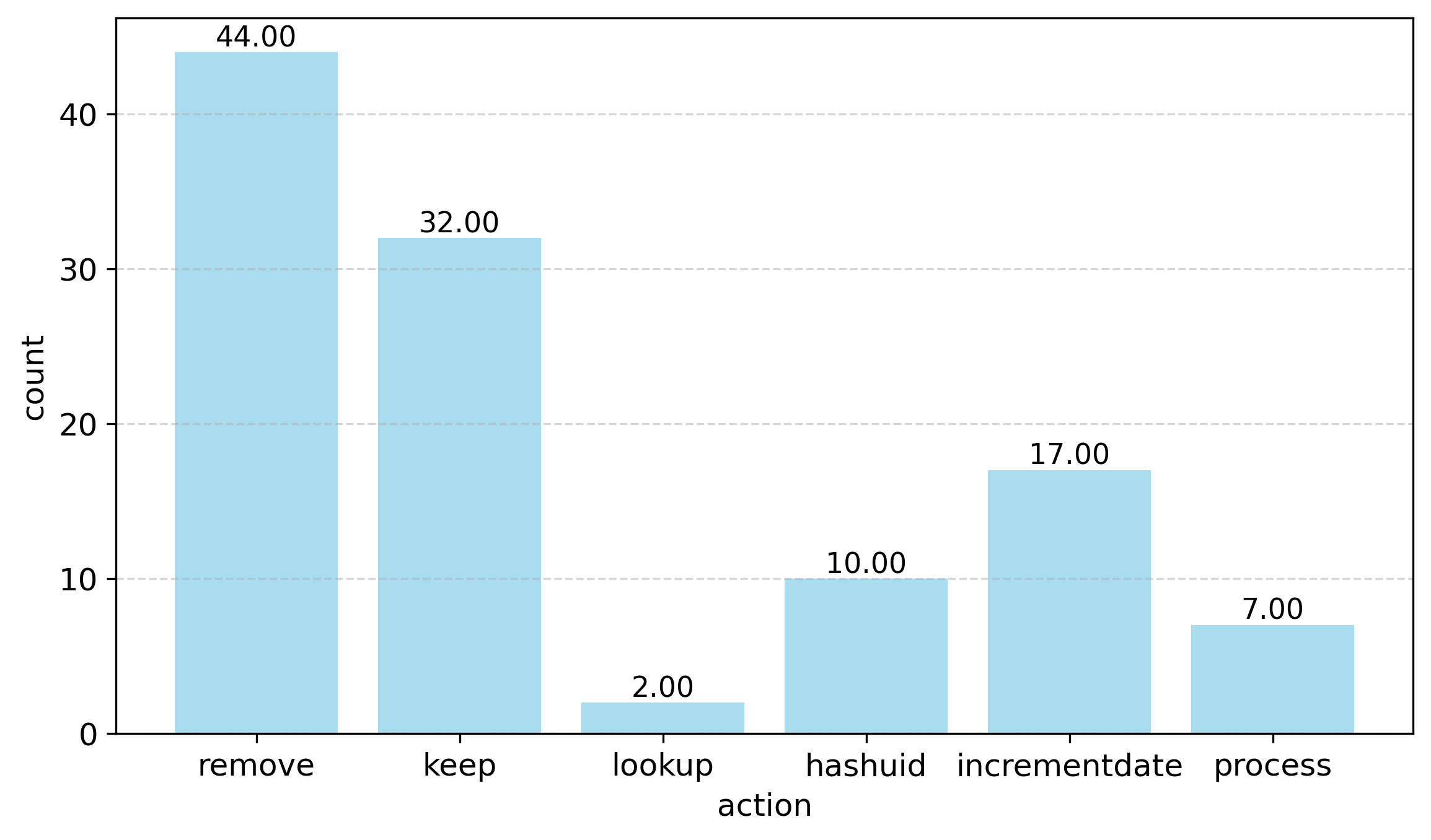}
    \caption{Distribution of action types in the validation dataset.}
\end{figure}
\subsection{Submission Results} According to the final leaderboard of the MIDI-B challenge, the latest version of our solution algorithm correctly executed 99.92\% of the required actions and ranked 2nd out of 10 teams that completed the challenge (from a total of 22 registered teams), as shown in the leaderboard. The total running time is about 41 hours and 41 minutes, and the pixel processing takes 99\% of the time (i.e., 41 hours and 11 minutes). The running time for each DICOM image file is 5.06 seconds. The winning team correctly executed 99.93\% of required actions. In other words, the winning team correctly executed around 58 more required actions than our team did.

\paragraph{Action Report.}The action report provides counts of both passed and failed values for each action, as shown in Table 8.

\begin{table}[h]
    \centering
    \caption{Action report detailing our solution’s performance on the test dataset.}
    \begin{tabular}{p{0.06 \columnwidth} p{0.38 \columnwidth}p{0.1 \columnwidth}p{0.15 \columnwidth} p{0.1\columnwidth}}
        \textbf{ } &\textbf{action} &\textbf{Fail} &\textbf{Pass} &\textbf{Total}\\
        \hline
        0 & \textless date\_shifted\textgreater & 2 & 2304 & 2306 \\
        1 & \textless patid\_consistent\textgreater & 0 & 429 & 429 \\
        2 & \textless pixels\_hidden\textgreater	& 0	& 15 & 15 \\
        3 &	\textless pixels\_retained\textgreater & 0 &	29471 &	29471 \\
        4 & \textless tag\_retained\textgreater & 8	& 121682 & 121690 \\
        5 & \textless text\_notnull\textgreater & 12 & 85311 & 85323 \\
        6 & \textless text\_removed\textgreater & 326 & 5490 & 5816 \\
        7 & \textless text\_retained\textgreater	& 131 & 254818 & 254949 \\
        8 & \textless uid\_changed\textgreater & 4 & 40629 & 40633 \\
        9 & \textless uid\_consistent\textgreater & 4 & 40629 & 40633\\
        Total & & 487 & 580778 & 581265
    \end{tabular}
\end{table}
\paragraph{Category Report.} The category report provides counts of both passed and failed values for each answer category, as shown in Table 9.

\begin{table}[h]
    \centering
    \caption{Category report detailing our solution’s performance on the test dataset.}
    \begin{tabular}{p{0.045 \columnwidth} p{0.14 \columnwidth}p{0.4 \columnwidth}p{0.1 \columnwidth} p{0.115\columnwidth}}
        \textbf{ } &\textbf{category} &\textbf{subcategory} &\textbf{Fail} &\textbf{Pass}\\
        \hline
        0 &  dicom & DICOM-IOD-1 & 20 & 170626 \\
        1 &  dicom & DICOM-IOD-2 & 0 & 36367 \\
        2 &  dicom	& DICOM-P15-BASIC-C	& 0 & 429 \\
        3 &	 dicom & DICOM-P15-BASIC-U & 4 & 40629 \\
        4 &  hipaa & HIPAA-A & 0 & 676 \\
        5 &  hipaa & HIPAA-B & 2 & 30 \\
        6 &  hipaa & HIPAA-C & 2 & 2816 \\
        7 &  hipaa	& HIPAA-D & 0 & 20 \\
        8 &  hipaa & HIPAA-G & 0 & 149 \\
        9 &  hipaa & HIPAA-H & 1 & 657\\
        10 &  hipaa & HIPAA-R & 4 & 41574\\
        11 &  tcia & TCIA-P15-BASIC-D & 0& 198\\
        12 &  tcia & TCIA-P15-BASIC-X & 0 & 13\\
        13 &  tcia & TCIA-P15-BASIC-X/Z/D & 0 & 147\\
        14 &  tcia & TCIA-P15-BASIC-Z & 0 & 482\\
        15 &  tcia & TCIA-P15-BASIC-Z/D & 0 & 11\\
        16 &  tcia & TCIA-P15-DESC-C & 8 & 2340\\
        17 &  tcia & TCIA-P15-DEV-C & 0 & 17\\
        18 &  tcia & TCIA-P15-DEV-K & 0 & 179\\
        19 &  tcia & TCIA-P15-MOD-C & 0 & 2565\\
        20 &  tcia & TCIA-P15-PAT-K & 0 & 1113\\
        21 &  tcia & TCIA-P15-PIX-K & 0 & 29471\\
        22 &  tcia & TCIA-PTKB-K & 117 & 31670\\
        23 &  tcia & TCIA-PTKB-X & 257 & 901\\
        24 &  tcia & TCIA-REV & 72 & 217698\\
        Total & &  487& 580778 & 581265
    \end{tabular}
\end{table}
\paragraph{Scoring Report.} The scoring report takes the categories from the previous report and assigns them to scoring categories, as shown in Table 10. The score is calculated in terms of the ratio of fail/pass actions.

\begin{table}[h]
    \centering
    \caption{Scoring report for our solution on the test dataset.}
    \begin{tabular}{p{0.2 \columnwidth} p{0.1 \columnwidth}p{0.15 \columnwidth}p{0.2 \columnwidth} p{0.115\columnwidth}}
        \textbf{Category} &\textbf{Fail} &\textbf{Pass} &\textbf{Total} &\textbf{Score}\\
        \hline
        All & 487 & 580,778 & 581,265 & 99.92\% \\
        \end{tabular}
\end{table}

\begin{figure}[h]
    \centering
    \includegraphics[width=0.85\linewidth]{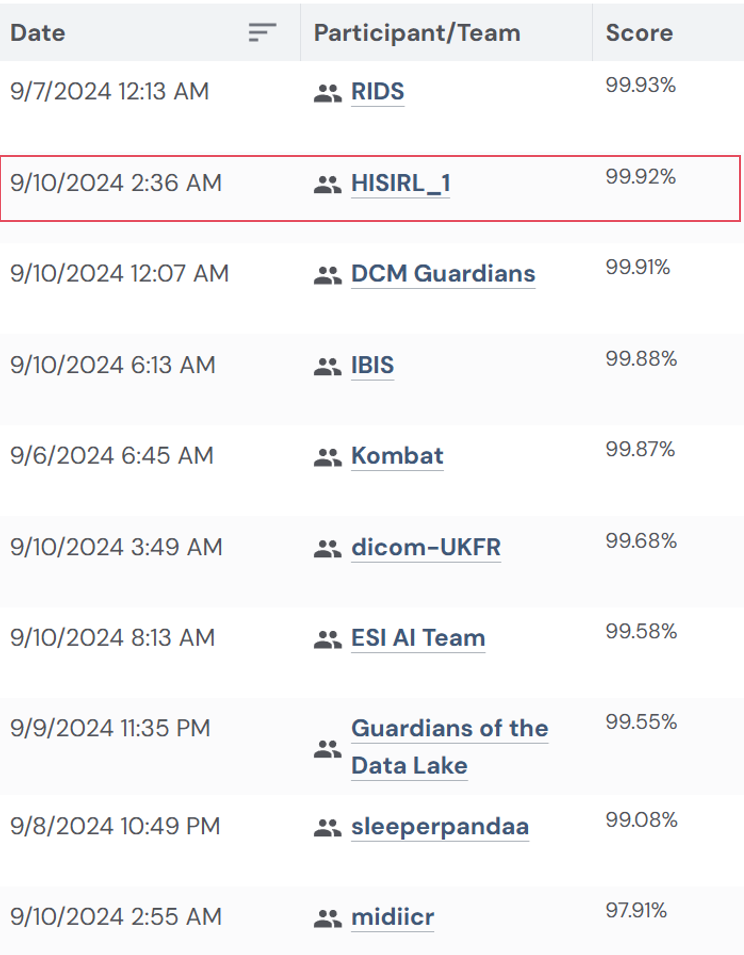}
    \caption{The screenshot of the challenge website showing the final scores.}
\end{figure}

\section{Discussion}
In conclusion, the evaluation results of our solution algorithm demonstrate the effectiveness of DICOM image de-identification in protecting patient privacy while preserving the data's utility in an automated manner. However, both our solution and the challenge have limitations that need to be addressed and improved upon in the future.
\subsection{Limitations of Our Algorithm}
Our proposed method performs well on specific datasets, but it has certain limitations in terms of generalizability. This means that when being applied to other datasets, it may not achieve the same results and might even fail to meet the expected standards \citep{robinson2014}. Therefore, the applicability of the method could be restricted, requiring further optimization and adjustment to ensure its effectiveness across a broader range of datasets \citep{bennett2018}.

While we have achieved some degree of effectiveness in removing sensitive information from text, there is still room for improvement in terms of accuracy. The current algorithm may either miss certain crucial sensitive data or incorrectly remove non-sensitive data. This can result in suboptimal outcomes that do not fully meet the expected security standards. Therefore, further refinement of our techniques for removing sensitive information, with a focus on enhancing accuracy, remains a key area of focus to ensure the reliability and precision of the results.
\subsection{Limitations of the MIDI-B Challenge}
\paragraph{Limitations regarding the dataset. } The dataset selected for the MIDI-B De-identification challenge is both novel and unique compared to other medical imaging datasets, such as those found in TCIA. It provides robust support for addressing critical issues in the de-identification of medical data, particularly in terms of patient privacy protection and data sharing \citep{rutherford2021a,rutherford2021b}.

We propose the following considerations to enhance its utility: The dataset should include detailed metadata and comprehensive documentation to enable researchers to understand and replicate the de-identification process. Additionally, the dataset's size and diversity should be increased to better support the training and validation of machine learning models, facilitating broader application in various de-identification scenarios.

\paragraph{Limitations regarding the privacy protection requirement.} The MIDI-B challenge currently requires only de- identification and pseudonymization, without incorporating anonymization considering an adversarial model. It is recommended that anonymization be included in the privacy protection requirements, particularly when handling sensitive data. Furthermore, the competition can introduce a re-identification attack model \citep{wan2015}, considering the potential for attackers to re-identify de-identified and pseudonymized data through external data sources or cross-referencing.



\acks{We acknowledge the organizers of Medical Image De-Identification Benchmark (MIDI-B) challenge 2024.}

%
\ethics{The work follows appropriate ethical standards in conducting research and writing the manuscript, following all applicable laws and regulations regarding treatment of animals or human subjects.}

\coi{The authors declare no conflicts of interest.}

\data{The data supporting the findings of this study are available at: \url{https://www.synapse.org/Synapse:syn53065760/wiki/627887}
}

\bibliography{sample}

\end{document}